\documentclass[a4paper,fleqn]{cas-dc}

\usepackage[numbers,sort&compress]{natbib}
\usepackage{url}

\newcommand{\method}{CHEAR}
\newcommand{\best}[1]{\textbf{#1}}
\newcommand{\conceptbox}[1]{\par\noindent #1\par}

\begin{document}
\let\WriteBookmarks\relax
\shorttitle{CHEAR for Human Activity Recognition}
\shortauthors{Lu et al.}
\title[mode=title]{CHEAR: Confusion-Hierarchy Evidence-Aware Reasoning for Cost-Efficient Human Activity Recognition}

\author[1]{Xiangqing Lu}
\ead{xqlu@stu.usc.edu.cn}
\author[1]{Tao Zhu}
\cormark[1]
\ead{tzhu@usc.edu.cn}
\author[1]{Mingxing Nie}
\ead{niemx@usc.edu.cn}
\author[2]{Liming Chen}
\ead{limingchen0922@dlut.edu.cn}
\affiliation[1]{organization={School of Computer, University of South China},
                city={Hengyang},
                country={China}}
\affiliation[2]{organization={School of Computer Science and Technology, Dalian University of Technology},
                city={Dalian},
                country={China}}
\cortext[1]{Corresponding author}

\begin{abstract}
Large language models (LLMs) offer a new approach to wearable-sensor-based human activity recognition. Retrieval-augmented generation (RAG) further introduces similar motion examples and discriminative sensor knowledge, enabling decisions grounded in observations relevant to the current query. More evidence, however, does not necessarily mean better-organized evidence. When online prompts and LLM agents primarily orchestrate inference, decisions about which activities to compare and when to invoke discriminative knowledge remain concentrated at inference time. As the activity vocabulary grows, the model must repeatedly process more candidates and associated knowledge, whereas pruning candidates too early risks excluding the true activity. We propose confusion-hierarchy evidence-aware reasoning (CHEAR), which moves part of this knowledge organization offline. CHEAR constructs and validates an activity hierarchy from recurrent cross-subject retrieval errors and builds corresponding activity-pair knowledge. At inference time, the hierarchy orders candidate comparisons; the LLM retains candidates progressively, and local discriminative knowledge is accessed only for the remaining activities. Across eight datasets, CHEAR achieves 86.79\% mean cross-subject accuracy, a 75.91\% candidate reduction rate, and 96.44\% true-activity retention. Per-sample token use is 85.15\% lower on average than in a variant without the hierarchical structure. Mean accuracy across eight cross-dataset transfer directions is 88.81\%. These results show that organizing candidates and discriminative knowledge around observed activity confusions reduces online LLM input while maintaining recognition performance.
\end{abstract}

\begin{graphicalabstract}
\includegraphics[width=\textwidth]{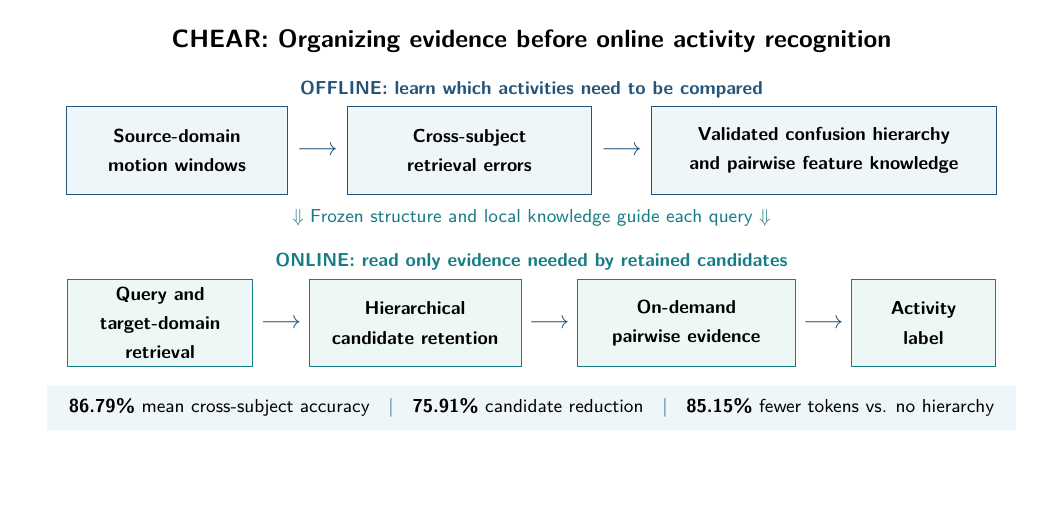}
\end{graphicalabstract}

\begin{highlights}
\item Confusion-guided organization of motion evidence for activity recognition.
\item Offline construction of activity hierarchies and local discriminative knowledge.
\item On-demand evidence use within hierarchically retained activity candidates.
\item Accurate recognition with fewer candidates and lower language-model token use.
\end{highlights}


\maketitle

\section{Introduction}

Wearable-sensor-based human activity recognition (HAR) maps multichannel motion signals to activity labels for applications such as health monitoring, sports analytics, and mobile interaction. Recent uses of LLMs in HAR have moved from direct interpretation of sensor sequences toward evidence-grounded decisions. HARGPT~\cite{hargpt} prompts an LLM with sensor sequences and activity descriptions. RAG-HAR~\cite{raghar} retrieves similar sensor windows and statistical descriptors as examples for the current query. ZARA~\cite{zara} combines position-specific motion retrieval with sensor knowledge in an LLM-based reasoning agent. Together, these studies show that observed motion patterns and signal knowledge can enter the decision context directly, reducing reliance on an LLM's interpretation of raw numerical sequences alone.

Retrieval augmentation addresses where decision evidence comes from, but adding evidence does not by itself organize it effectively. Similar examples are generally returned by activity, whereas local features describe distinctions between particular pairs of activities. Their usefulness depends on which activities must currently be distinguished and when the corresponding features are consulted. For example, walking versus ascending stairs involves motion rhythm and vertical-axis changes; sitting versus standing depends more on body posture; and different eating activities may share hand movement while differing in local wrist dynamics. In pipelines orchestrated primarily by online prompts and LLM agents, these comparison relationships, candidate-screening order, and knowledge-access decisions are still largely determined after each query arrives. More activity classes therefore entail more candidates and related knowledge to process; removing candidates prematurely may eliminate the true activity. This reliance on online organization also blurs the division between knowledge processing that could be completed in advance and activity decisions that must remain online, complicating future edge--cloud task allocation.

We therefore study not only how to retrieve similar examples but also how observed activity confusions can organize candidate comparisons in advance and bring discriminative knowledge into inference only when needed. We propose \method\ (confusion-hierarchy evidence-aware reasoning). Offline, we measure the direction and frequency of errors made by leave-one-subject-out retrieval in the source domain, construct an activity-confusion hierarchy, select groups using an independent source validation set, and build local feature knowledge for validated activity pairs. Online, the LLM uses current retrieval evidence to retain a variable number of candidates at each level of the hierarchy. Pairwise knowledge is accessed only when both relevant activities remain, supporting local comparison and final label selection. Stable activity relationships are thus supplied by an offline structure, while the LLM makes choices from evidence for the current query without having to organize every activity and its discriminative knowledge on its own.

Across eight public datasets, \method\ achieves 86.79\% mean cross-subject accuracy and 86.89\% macro-F1. The final candidate set contains 2.38 activities on average, corresponding to a 75.91\% candidate reduction rate and 96.44\% true-activity retention. Relative to the variant without the hierarchical structure, average per-sample token use decreases by 85.15\%. Mean accuracy across eight cross-dataset transfer directions is 88.81\%. These results evaluate the proposed organization of activity relationships and discriminative knowledge in terms of classification, candidate retention, and model input.

The principal contribution is to turn reproducible source-domain activity confusions from a post-hoc error analysis into an offline candidate structure that determines the order of online comparisons and the scope of local knowledge access. The framework retains the LLM's ability to reason over query-specific evidence while reducing the number of candidates and the amount of knowledge it processes online. The separation between offline knowledge organization and online activity decisions also provides a basis for future work on allocating tasks between edge devices and the cloud.
\section{Related Work}
Research on wearable-sensor HAR has addressed transferable representation learning from continuous signals, the alignment of motion signals with activity semantics, and the use of external evidence for individual queries. MOMENT~\cite{moment} pretrains on time series from diverse sources to provide general representations for downstream tasks, including classification. Mantis~\cite{mantis} studies lightweight representations for time-series classification through contrastive pretraining on synthetic data. Scaling Wearable Foundation Models~\cite{scalingwearable} examines the effects of data, model, and compute scale across everyday wearable signals and evaluates pretrained representations for motion and activity recognition. These studies provide encoders for sensor windows, while the relationship between motion signals and activity descriptions requires additional modeling.

Cross-modal studies introduce semantic information into motion representations through images, text, and sensor context. ImageBind~\cite{imagebind} uses images as a bridge to map six modalities, including inertial measurements and text, into a shared representation. IMU2CLIP~\cite{imu2clip} aligns inertial motion sequences with egocentric video and text, enabling cross-modal retrieval and activity recognition. To account for changes in body placement and device orientation, UniMTS~\cite{unimts} generates motion signals from skeletal actions and aligns them with text to learn activity semantics. ZeroHAR~\cite{zerohar} incorporates sensor type, coordinate axis, and wearing position into motion representations for recognizing activities unseen during training. IMUGPT~\cite{imugpt} follows another cross-modal route: it generates activity descriptions, synthesizes three-dimensional motion, and converts it to virtual sensor sequences for training activity classifiers.

Language models have also been used to interpret sensor inputs directly, rather than only to learn motion representations or generate training data. HARGPT~\cite{hargpt} examines whether an LLM can recognize activities from inertial signals and activity descriptions supplied through prompts. SensorLLM~\cite{sensorllm} first aligns sensor signals with text describing their trends and then trains for HAR. SensorLM~\cite{sensorlm} constructs training data through hierarchically generated sensor captions and learns a joint sensor--language representation. LLM4HAR~\cite{llm4har} adapts pretrained language models through sensor embeddings and sensor knowledge, using partial training and model compression to support on-device inference. These approaches address prompting, model alignment, and model efficiency, respectively.

When a decision requires examples relevant to the current sample, retrieval-augmented methods bring similar labeled motion windows into inference. RAG-HAR~\cite{raghar} retrieves similar windows using sensor statistical descriptors and passes the results to a language model for classification. RAG-HAR+~\cite{ragharplus} selects dataset-specific retrieval features offline, classifies samples with sufficient retrieval evidence by nearest-neighbor voting, and invokes an LLM only for uncertain samples. ZARA~\cite{zara} combines motion-example retrieval with discriminative sensor knowledge, allowing a language-model agent to select cues and reason progressively over candidate activities. From general representations and cross-modal semantics to retrieved evidence, these studies expand the information available to HAR and motivate treating the organization of candidates and discriminative knowledge as a separate problem.

\section{Problem Definition and Design Principles}
Let $\mathcal{Y}=\{1,\ldots,K\}$ be the activity set. A query window $x$ contains inertial-sensor sequences from one or more body locations. Source-domain data are separated by purpose. Within the construction pool, subjects rotate between query and reference roles in leave-one-subject-out folds: all windows of the held-out subject are queries, and windows of the other construction subjects form the reference set. An independent source validation set is used only to select the hierarchy, confirm activity pairs, and choose feature budgets. Target test labels are not used for hierarchy construction, feature selection, or online decisions; target test windows enter as unlabeled queries only after these components have been frozen.

The system must output one label $\hat y\in\mathcal{Y}$. In addition to accuracy and macro-F1, we record the mean size of the final leaf-activity candidate set $S(x)$. For a fixed dataset of $N$ queries, candidate reduction is $1-\frac{1}{N}\sum_{n=1}^{N}|S(x_n)|/K$, and true-activity retention is $\frac{1}{N}\sum_{n=1}^{N}\mathrm{I}[y_n\in S(x_n)]$. Candidate-set metrics measure whether routing preserves an opportunity for subsequent decisions; they do not replace final classification accuracy.

\conceptbox{\textbf{Meaning of the activity-confusion hierarchy.} In practical terms, it places activities that retrieval repeatedly confuses into the same candidate group. Its inputs are directed confusion matrices from source construction queries and an independent validation set. A null model first removes the confusion expected from overall class difficulty. Candidate groups are then evaluated by their modularity gain, and only groups with strictly positive gains in both datasets are retained. Online, the model selects a nonempty subset of current candidate units. Unselected groups and their activities are removed; selected groups expand by one level, while singleton activities remain unchanged. The outputs are root candidate units, their within-group children, and confirmed activity pairs. For example, if walking and ascending stairs exhibit above-expected confusion in both directions in both source evidence sets, they may first appear as one group and expand into two activities only after that group is retained.}

The design follows two principles. First, \textbf{the hierarchy is derived only from reproducible source-domain evidence}; neither an LLM nor a person modifies it. Second, \textbf{local evidence applies only to its registered endpoints}: an activity-pair report compares its two specified activities and cannot become a global vote over other candidates.

\section{Method}
\subsection{Overview}
\begin{figure*}[t]
\centering
\includegraphics[width=\textwidth]{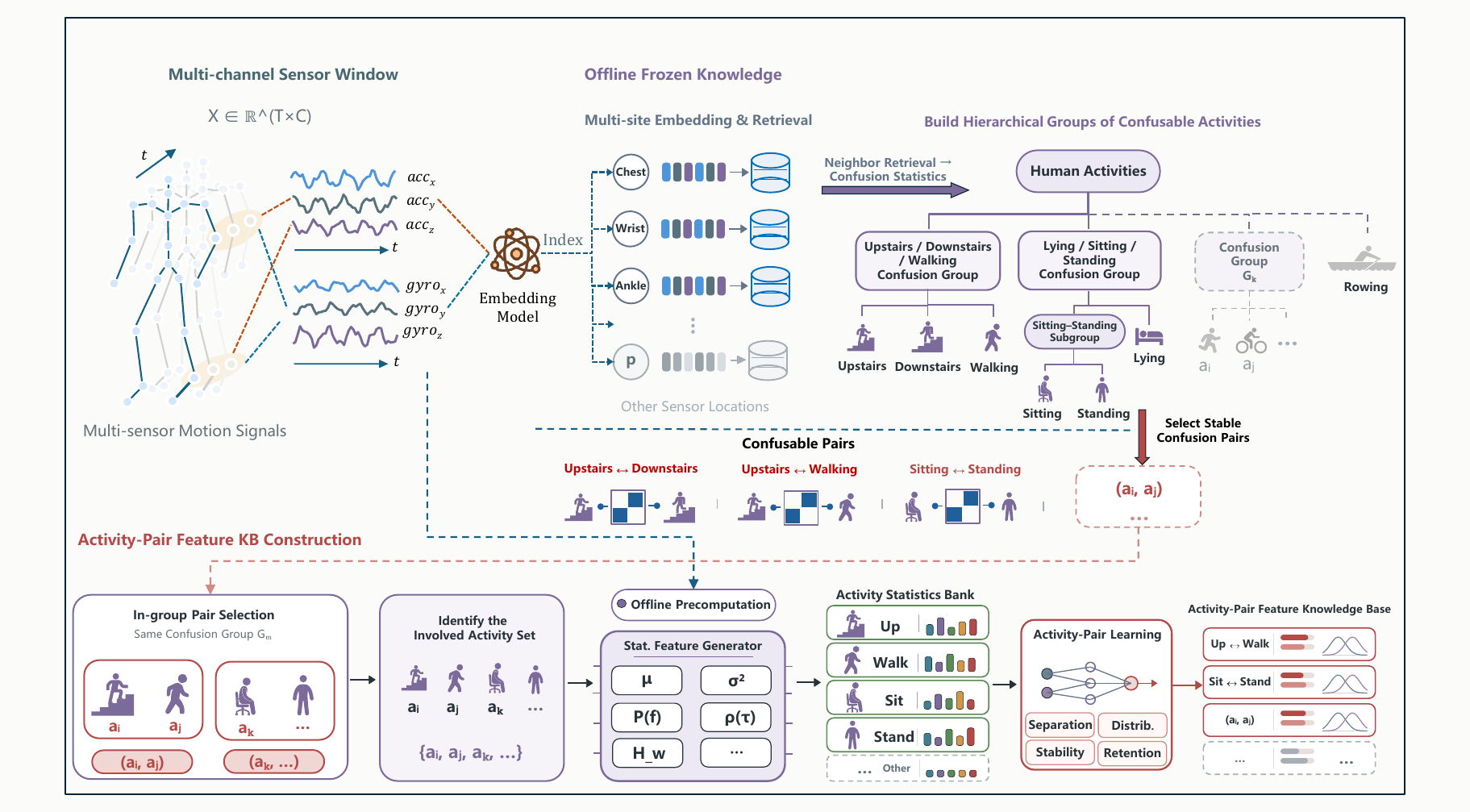}
\caption{Offline knowledge construction in \method.}
\label{fig:offline-overview}
\end{figure*}

\begin{figure*}[t]
\centering
\includegraphics[width=\textwidth]{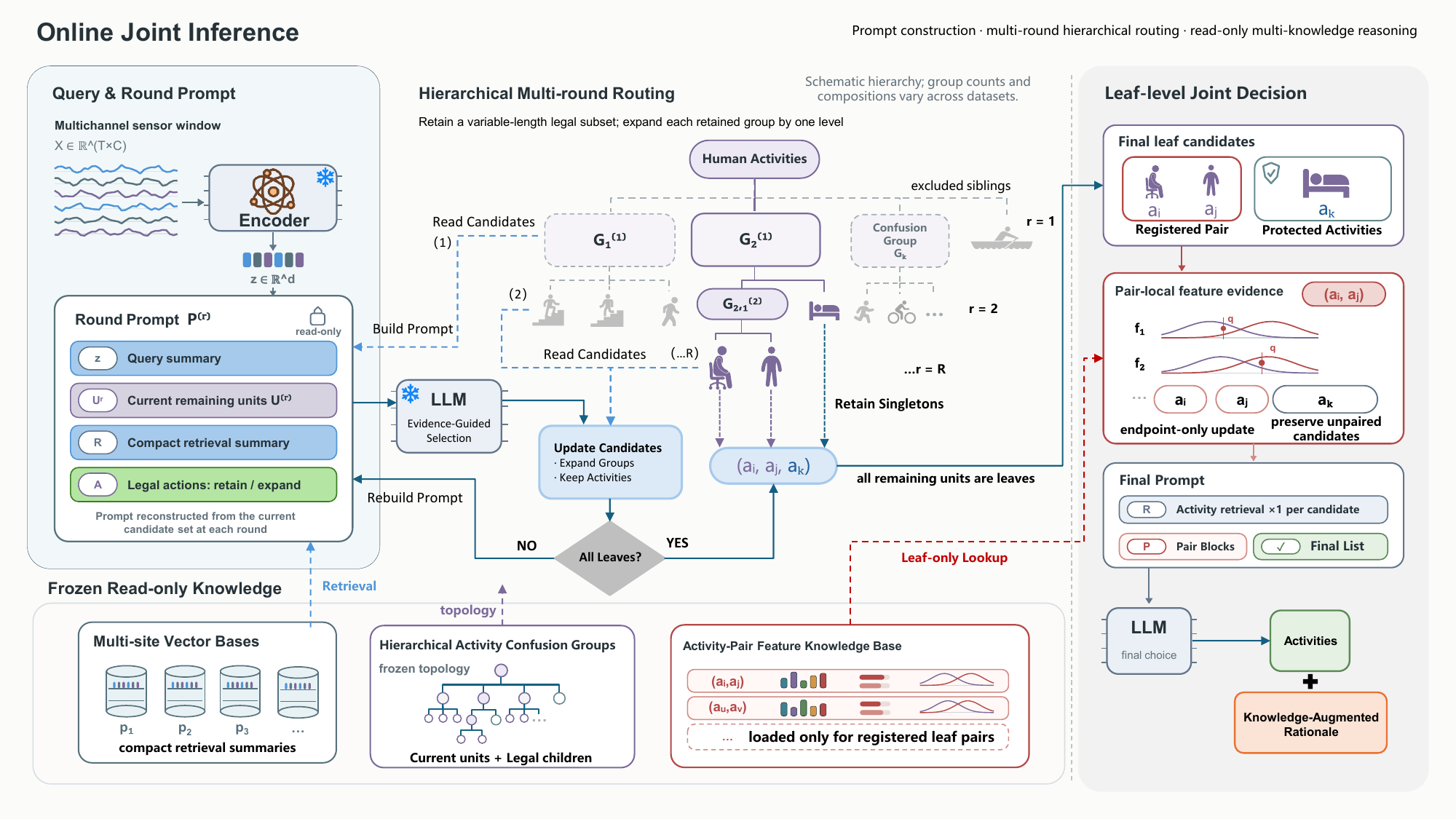}
\caption{Online hierarchical inference in \method.}
\label{fig:online-overview}
\end{figure*}

\textbf{Offline knowledge construction.} Fig.~\ref{fig:offline-overview} summarizes the preparation completed before testing. The input consists of labeled, multi-position inertial-sensor windows from the source domain. A frozen encoder produces retrieval vectors for each sensor position; position-specific reference stores are then used to measure activity confusion under leave-one-subject-out retrieval. Candidate groups are generated from the directed confusion matrix and filtered using an independent source validation set. For difficult activity pairs confirmed by the hierarchy, the system computes predefined statistical features from the corresponding reference windows, trains activity-pair models, and selects feature budgets using independent source validation. The stage produces three frozen outputs: position-specific retrieval stores, the activity-confusion hierarchy, and local feature knowledge for registered activity pairs.

\textbf{Online hierarchical inference.} Fig.~\ref{fig:online-overview} shows inference for one query window. The same frozen encoder processes the query, and a compact retrieval summary is obtained from the target dataset's position-specific reference stores. Inference starts from root candidate units. Given the current candidates and retrieval evidence, the LLM retains a variable-size subset. Each retained confusion group expands by one level, whereas a singleton activity proceeds unchanged. This continues until every remaining candidate is an individual activity. The system then reads local feature reports only for registered pairs whose activities both remain among the candidates; unpaired or unregistered activities continue to participate in the final decision based on retrieval evidence. The output is one activity label from the remaining candidates. Offline preparation determines the available hierarchy and local knowledge, while online inference accesses only the portions relevant to the query and remaining candidates rather than comparing every activity and every pairwise feature at once.

\subsection{Frozen Encoding and Position-Specific Retrieval}
\textbf{Input and processing.} A query window from each sensor position is linearly resampled to 512 time points and passed to a frozen Mantis-8M encoder. The encoder produces a 256-dimensional vector per channel, followed by $\ell_2$ normalization. Reference vectors are indexed by sensor position and activity using FAISS IndexFlatIP\cite{faiss}. For each activity and position, at most the 100 most similar reference items are retained; the globally top 100 results are also retained per position.

\textbf{Activity scores.} The similarity between normalized query vector $u$ and reference vector $v$ is
\begin{equation}
 s(u,v)=u^{\top}v=\cos(u,v).
\end{equation}
An activity score is the sum of raw cosine similarities across positions. It ranks retrieval results and forms part of the compact evidence description; it is neither interpreted as a probability nor used directly as a group confidence score. Reciprocal rank fusion\cite{rrf} is used only to select reference examples for activity pairs: an example ranked $r\geq1$ in a list contributes $1/(60+r)$, with at most 100 items from each list.

\textbf{Retrieval evidence.} For each eligible activity, the module summarizes hit counts, similarity quality, and results across sensor positions. Online inference reads these compact summaries to compare candidate units at the current hierarchy level.

\subsection{Constructing a Hierarchy from Directed Confusion}
\subsubsection{Confusion Matrix and Null Model}
We perform leave-one-subject-out retrieval among construction subjects and retain only off-diagonal errors. The conditional probability that the retrieval rule incorrectly predicts $j$ when the true activity is $i$ is
\begin{equation}
 A_{ij}=P(\hat y=j\mid y=i),\qquad A_{ii}=0.
\end{equation}
Both $i\rightarrow j$ and $j\rightarrow i$ are retained. Let
\begin{equation}
 k_i^{\mathrm{out}}=\sum_j A_{ij},\quad
 k_j^{\mathrm{in}}=\sum_i A_{ij},\quad
 m=\sum_{i,j}A_{ij},
\end{equation}
The excess confusion after accounting for overall activity difficulty is
\begin{equation}
 B_{ij}=A_{ij}-\frac{k_i^{\mathrm{out}}k_j^{\mathrm{in}}}{m}.
\end{equation}
$B_{ij}>0$ means that the observed $i\rightarrow j$ confusion exceeds the expectation of a null model preserving in- and out-strengths; it does not merely indicate that an error occurred.

\subsubsection{Candidate Merging and Independent Validation}
Given an activity partition $c$, directed modularity\cite{leichtnewman} is
\begin{equation}
 Q(c)=\frac{1}{m}\sum_{i,j}B_{ij}\,\delta(c_i,c_j).
\end{equation}
Starting from singleton activities, the method repeatedly merges the two groups with the largest strictly positive modularity gain and stops when no positive gain remains.

The construction set generates candidate merge paths. The independent source validation set may select only candidates on these paths whose gain over the all-singleton partition is strictly positive; it cannot introduce a structure absent from construction. The same procedure is then applied recursively to the induced submatrix of each accepted group. Thus, the method imposes neither a fixed depth nor a binary-tree constraint, and activities without stable confusion are not forced into artificial intermediate categories.

\subsubsection{Registering Difficult Activity Pairs}
Bidirectional excess confusion between specific activities $i$ and $j$ is
\begin{equation}
 E(i,j)=B_{ij}+B_{ji}.
\end{equation}
An activity pair is registered only if $E(i,j)>0$ in both the construction and independent source validation sets and both classes have validation examples. A within-group pair is attached to the deepest confusion group containing both endpoints. A cross-group pair is stored as a root-level lateral relationship without changing the main hierarchy.

\conceptbox{\textbf{Meaning of activity-pair feature knowledge.} This is a description of sensor differences between two activities that are easily confused. During construction, the method reads source-domain acceleration, angular velocity, and retrieved reference examples for a registered pair; trains candidate tabular models; computes permutation importance; and selects the number of feature families on the independent source validation set. It outputs a pair identifier, available features, source statistics, and frozen reliability. During online inference, the corresponding report is loaded only if both endpoints remain among the current candidates. The report can compare its left and right endpoints but cannot affect a third activity. For example, a walking--ascending-stairs report provides local evidence only for those two activities and cannot be used to remove standing.}

\subsection{Local Feature Knowledge for Activity Pairs}
\textbf{Inputs.} Feature knowledge is built only for registered activity pairs. Candidate feature families include time-domain and frequency-domain statistics, short-time Fourier transforms, autocorrelation, impact rate of change, fourth-order autoregression, recurrence rate, wavelet packets, permutation entropy, gravity direction, and inter-axis correlation.

\textbf{Models and importance.} Leave-one-subject-out evaluation is performed on construction subjects. In each fold, AutoGluon\cite{autogluon} fits candidate tabular models on the training subjects: GBM (gradient boosting machine), XGB (extreme gradient boosting), CAT (CatBoost), RF (random forest), XT (extremely randomized trees), and LR (logistic regression). Macro-F1 and permutation importance over five repetitions are computed on the held-out subject. Negative importance values are clipped to zero, and stable ties are resolved by feature identifier. Positive-class weights are restricted to $[1,50]$ so that majority-class examples do not dominate minority-class activity-pair training.

\textbf{Budget selection.} Candidate feature-family budgets are $M\in\{4,8,12\}$. The independent source validation set selects $M$ by activity-pair macro-F1, choosing the smaller budget in a tie. Feature knowledge is enabled online only if the best budget strictly outperforms a similarity-only activity-pair baseline; all other pairs use retrieval evidence alone.

\textbf{Online use.} The tabular models above are used only offline for feature screening and budget selection; they are not called for online queries. For a query, the selected feature value $x$ is computed from the target window and compared with frozen source statistics $\mu_c,s_c$ of candidate activity $c$:
\begin{equation}
 z=\frac{x-\mu_c}{\max(s_c,10^{-6})}.
\end{equation}
A smaller $|z|$ indicates that the query value is closer to the activity's source statistics; it is not itself a programmatic activity decision. The system summarizes distances and frozen reliability for both endpoints by feature family and supplies them, together with endpoint-specific retrieval summaries, to the LLM. The model judges whether the left endpoint, right endpoint, or neither can be selected for that pair. If the pair does not meet the feature-enablement criterion or required features are unavailable, no feature report is loaded and the remaining candidates are judged from retrieval evidence.

\subsection{Variable-Size Hierarchical Candidate Retention}
Online root candidates comprise top-level confusion groups and root-level singleton activities. Let $U_t$ be the candidate units at round $t$. The LLM reads each unit's identifier and compact retrieval summary and outputs a variable-size set of retained units $R_t\subseteq U_t$. Candidates are updated by
\begin{equation}
U_{t+1}=\bigcup_{u\in R_t}\begin{cases}\mathrm{children}(u),&u\text{ is a confusion group},\\\{u\},&u\text{ is a singleton activity}.\end{cases}
\end{equation}
Thus, the system neither fixes a top-$N$ candidate count nor requires inference to follow a single path:
\begin{enumerate}
  \item A retained confusion group expands by one level; an unretained group and its activities leave the candidate set.
  \item A retained singleton activity proceeds unchanged to the next round.
  \item If hit counts, similarity quality, and conclusions across sensor positions conflict, the model retains the smallest reasonable set covering the conflict rather than being forced onto one path.
  \item Registered and available activity pairs are enumerated only after every remaining candidate is a leaf activity.
\end{enumerate}

At the activity-pair stage, the model labels each available report as favoring the left endpoint, favoring the right endpoint, or inconclusive. Each report compares only its two endpoints. An inconclusive result provides no basis for deletion, and candidates without a direct pairwise comparison remain. A candidate is removed only when all its direct comparisons favor the opposing endpoint. The final prompt presents one global retrieval summary per remaining activity and appends local feature evidence for registered pairs. The model then selects one activity label from the remaining leaf activities.

\begin{figure*}[!t]
\centering
\includegraphics[width=\textwidth]{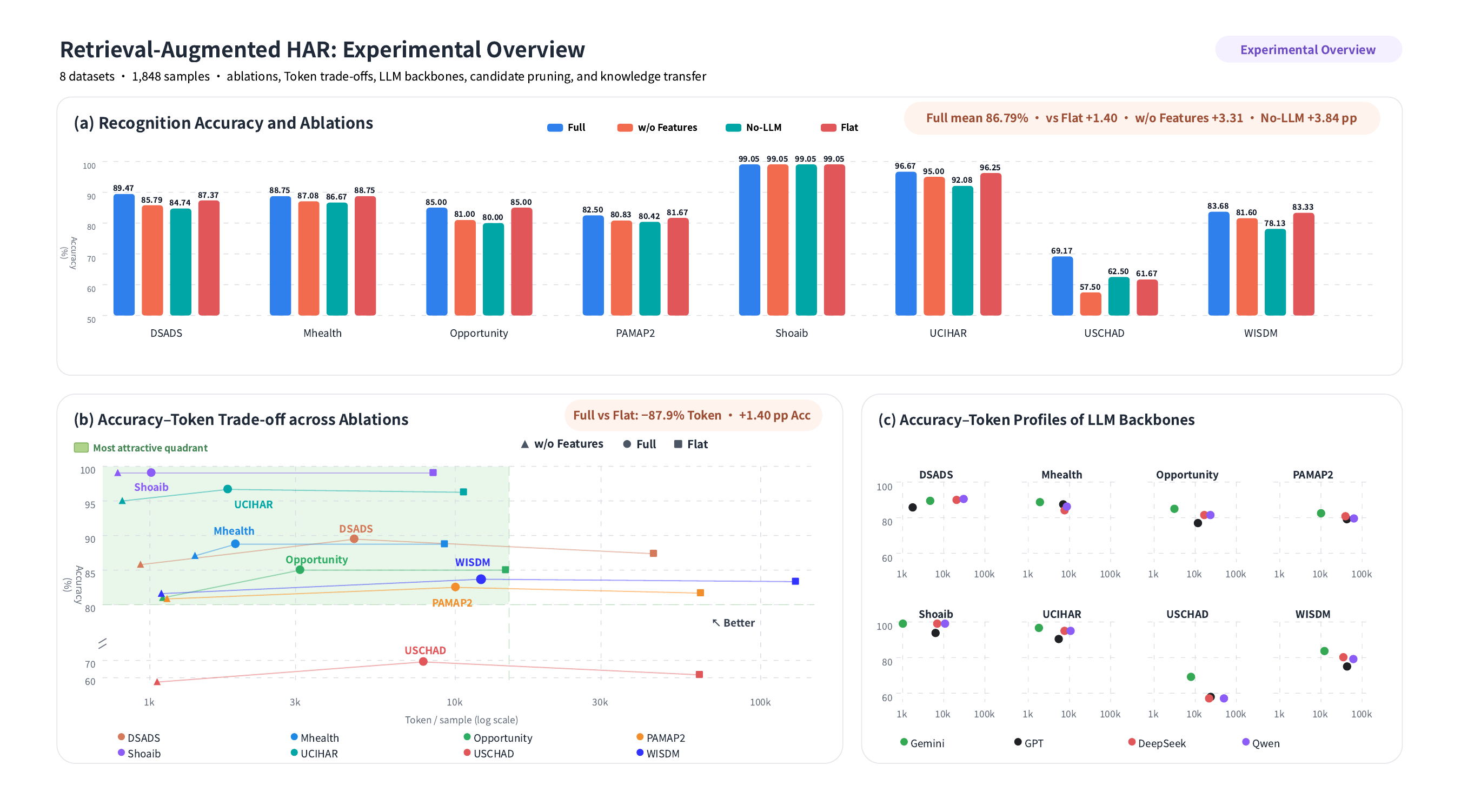}
\caption{Overview of cross-subject accuracy, token cost, and results across foundation models.}
\label{fig:experimental-overview}
\end{figure*}

\section{Experimental Setup}
\subsection{Datasets and Evaluation Protocol}
We evaluate on eight public motion datasets: DSADS\cite{dsads}, Mhealth\cite{mhealth}, Opportunity\cite{opportunity}, PAMAP2\cite{pamap2}, Shoaib\cite{shoaib}, UCIHAR\cite{ucihar}, USCHAD\cite{uschad}, and WISDM\cite{wisdm}. The retrieval encoder, indexing rules, hierarchy, activity-pair registry, and feature budgets are frozen before testing. Activities, sensor positions, subject splits, and window preprocessing are described in the appendix.

Cross-dataset experiments align only shared activities and comparable body locations. The transferred components are limited to the source-domain hierarchy and activity-pair feature knowledge. Online retrieval always uses a target-domain reference set and never retrieves raw source-domain windows. This separates the transferability of knowledge from direct reuse of source examples in the target domain.

\subsection{Comparators and Metrics}
Comparators fall into three categories: direct prompting, pretrained recognition, and retrieval-augmented reasoning. Direct prompting includes HARGPT and Gemini reading numerical sequences or sensor plots, as well as Gemini reading Markdown tables. Pretrained methods include NormWear\cite{normwear}, ImageBind, IMU2CLIP, and UniMTS, which recognize activities through pretrained representations or text matching; IMUGPT uses synthetic motion data to train a classifier. The retrieval-augmented comparator, ZARA, combines position-specific retrieval, sensor statistical knowledge, and LLM-agent reasoning.

The main configuration and cross-dataset experiments use Gemini 3 Flash Preview\cite{gemini_family} (\texttt{gemini-3-flash-preview}). The multi-model study also uses GPT-4.1 mini\cite{gpt4_report}, DeepSeek-V3.2\cite{deepseek_v32}, and Qwen3-Next-80B-A3B-Instruct\cite{qwen3_report}.

The main metrics are accuracy and macro-F1. Structural effects are also assessed by the number of root candidate units, mean final candidate count, candidate reduction, and true-activity retention. Cost is measured in tokens per sample and wall-clock latency. Cross-subject means are arithmetic means over the eight datasets. In the cross-dataset table, accuracy and macro-F1 are equally weighted across eight transfer directions; candidate count, reduction, and retention are aggregated over all query windows.

\section{Results}
Fig.~\ref{fig:experimental-overview} summarizes cross-subject accuracy, token cost, and results across foundation models; the corresponding tables give the exact values.

\begin{table*}[!t]
\centering
\caption{Cross-subject comparison with prior methods (accuracy and macro-F1, \%). Results for prior methods are taken from ZARA\cite{zara}. Means are arithmetic means across eight datasets. For each metric in each dataset and in the mean column, the highest value is bold and the second-highest is underlined; tied values receive the same marking.}
\label{tab:model-detail}
\rmfamily
{\renewcommand{\arraystretch}{0.86}\setlength{\tabcolsep}{2.0pt}
\resizebox{\textwidth}{!}{%
\begin{tabular}{@{}l*{18}{r}@{}}
\toprule
 & \multicolumn{6}{c}{Easy} & \multicolumn{6}{c}{Medium} & \multicolumn{4}{c}{Hard} & \multicolumn{2}{c}{Mean}\\
\cmidrule(lr){2-7}\cmidrule(lr){8-13}\cmidrule(lr){14-17}\cmidrule(l){18-19}
Model/Method & \multicolumn{2}{c}{Opportunity} & \multicolumn{2}{c}{UCIHAR} & \multicolumn{2}{c}{Shoaib} & \multicolumn{2}{c}{PAMAP2} & \multicolumn{2}{c}{USCHAD} & \multicolumn{2}{c}{Mhealth} & \multicolumn{2}{c}{WISDM} & \multicolumn{2}{c}{DSADS} & \multicolumn{2}{c}{Mean}\\
 & Acc & F1 & Acc & F1 & Acc & F1 & Acc & F1 & Acc & F1 & Acc & F1 & Acc & F1 & Acc & F1 & Acc & F1\\
\midrule
\# Classes & \multicolumn{2}{c}{4} & \multicolumn{2}{c}{6} & \multicolumn{2}{c}{7} & \multicolumn{2}{c}{12} & \multicolumn{2}{c}{12} & \multicolumn{2}{c}{12} & \multicolumn{2}{c}{18} & \multicolumn{2}{c}{19} & \multicolumn{2}{c}{--}\\
\# Channels & \multicolumn{2}{c}{30} & \multicolumn{2}{c}{6} & \multicolumn{2}{c}{30} & \multicolumn{2}{c}{18} & \multicolumn{2}{c}{6} & \multicolumn{2}{c}{15} & \multicolumn{2}{c}{6} & \multicolumn{2}{c}{30} & \multicolumn{2}{c}{--}\\
\specialrule{0.06em}{0.12em}{0em}\specialrule{0.06em}{0.08em}{0.12em}
HARGPT$_{\mathrm{Text}}$ & 21.0 & 19.2 & 29.6 & 17.4 & 27.1 & 19.2 & 12.1 & 6.2 & 13.8 & 7.3 & 12.1 & 6.0 & 5.6 & 1.8 & 10.5 & 6.2 & 16.5 & 10.4\\
Gemini$_{\mathrm{Text}}$ & 26.5 & 19.8 & 24.2 & 13.0 & 27.1 & 17.6 & 15.0 & 10.2 & 14.2 & 5.4 & 25.4 & 20.8 & 11.1 & 7.6 & 13.2 & 8.6 & 19.6 & 12.9\\
Gemini$_{\mathrm{Table}}$ & 29.0 & 22.3 & 21.3 & 9.8 & 27.6 & 18.7 & 11.7 & 7.2 & 17.1 & 9.7 & 22.9 & 18.3 & 10.1 & 7.8 & 16.3 & 10.3 & 19.5 & 13.0\\
HARGPT$_{\mathrm{Plot}}$ & 21.5 & 15.6 & 28.3 & 15.7 & 24.3 & 14.2 & 10.0 & 6.9 & 14.6 & 8.7 & 15.0 & 11.0 & 5.9 & 2.8 & 7.9 & 4.5 & 15.9 & 9.9\\
Gemini$_{\mathrm{Plot}}$ & 23.5 & 21.3 & 20.6 & 31.7 & 31.4 & 24.1 & 10.4 & 6.9 & 10.8 & 5.3 & 19.2 & 17.4 & 9.4 & 7.2 & 10.0 & 4.8 & 18.3 & 13.5\\
\midrule
NormWear & 23.0 & 23.8 & 17.9 & 11.4 & 15.2 & 11.7 & 9.2 & 2.7 & 10.0 & 5.8 & 8.3 & 2.2 & 4.2 & 1.4 & 3.7 & 2.2 & 11.4 & 7.7\\
IMUGPT & 38.5 & 28.7 & 32.5 & 21.6 & 26.7 & 15.2 & 12.9 & 3.8 & 2.9 & 1.9 & 8.3 & 2.8 & 5.9 & 2.1 & 7.4 & 3.6 & 16.9 & 10.0\\
ImageBind & 35.5 & 30.0 & 28.8 & 19.9 & 36.7 & 30.2 & 18.8 & 10.2 & 7.9 & 1.8 & 17.9 & 11.1 & 8.0 & 4.7 & 10.5 & 5.7 & 20.5 & 14.2\\
IMU2CLIP & 36.5 & 34.4 & 33.3 & 22.8 & 39.5 & 34.5 & 15.8 & 11.6 & 16.3 & 10.5 & 16.3 & 14.3 & 10.1 & 5.9 & 13.7 & 9.2 & 22.7 & 17.9\\
UniMTS & 33.5 & 24.8 & 37.1 & 23.9 & 51.9 & 40.6 & 32.9 & 29.2 & 29.6 & 24.2 & 65.4 & 58.8 & 30.2 & 28.5 & 34.7 & 27.0 & 39.4 & 32.1\\
\midrule
ZARA$_{\mathrm{Qwen\text{-}80B}}$ & 80.5 & 79.8 & 76.7 & 75.4 & 93.8 & 93.5 & 67.5 & 67.0 & 47.1 & 49.2 & 82.1 & 81.6 & 57.6 & 56.4 & 81.6 & 81.3 & 73.4 & 73.0\\
ZARA$_{\mathrm{GPT}}$ & \underline{86.5} & \underline{86.5} & 85.0 & 85.0 & 93.3 & 93.2 & 72.5 & 72.8 & 56.7 & 57.4 & 82.5 & 80.6 & 60.8 & 59.5 & 82.6 & 82.3 & 77.5 & 77.2\\
ZARA$_{\mathrm{Gemini}}$ & \best{92.5} & \best{92.5} & 90.0 & 90.0 & \underline{97.1} & \underline{97.1} & 76.7 & 76.9 & \underline{60.0} & \underline{60.1} & 86.3 & 86.1 & 65.6 & 64.1 & 84.2 & 84.4 & 81.6 & 81.4\\
\specialrule{0.06em}{0.12em}{0em}\specialrule{0.06em}{0.08em}{0.12em}
\method$_{\mathrm{GPT}}$ & 77.00 & 76.78 & 90.42 & 90.44 & 93.81 & 93.90 & 79.17 & 80.68 & 57.92 & 56.96 & \underline{87.50} & \underline{87.41} & 75.00 & 75.03 & 85.79 & 85.40 & 80.83 & 80.83\\
\method$_{\mathrm{DeepSeek}}$ & 81.50 & 81.37 & \underline{95.00} & \underline{94.93} & \best{99.05} & \best{99.05} & \underline{80.83} & \underline{82.34} & 57.08 & 57.33 & 84.17 & 83.85 & \underline{80.21} & \underline{79.67} & \underline{90.00} & \underline{89.66} & 83.48 & 83.53\\
\method$_{\mathrm{Qwen}}$ & 81.50 & 81.36 & \underline{95.00} & \underline{94.93} & \best{99.05} & \best{99.05} & 79.58 & 81.28 & 57.08 & 57.12 & 86.25 & 85.97 & 79.17 & 78.46 & \best{90.53} & \best{90.24} & \underline{83.52} & \underline{83.55}\\
\method$_{\mathrm{Gemini}}$ & 85.00 & 85.13 & \best{96.67} & \best{96.64} & \best{99.05} & \best{99.05} & \best{82.50} & \best{83.43} & \best{69.17} & \best{69.58} & \best{88.75} & \best{88.62} & \best{83.68} & \best{83.50} & 89.47 & 89.18 & \best{86.79} & \best{86.89}\\
\bottomrule
\end{tabular}}
}
\end{table*}

\subsection{Cross-Subject Main Experiment}
Table~\ref{tab:model-detail} orders the eight datasets by increasing number of activities. Among prior methods, ZARA with Gemini has the highest mean accuracy (81.6\%) and macro-F1 (81.4\%). \method\ with Gemini reaches 86.79\% and 86.89\%, exceeding those tabulated values by 5.19 and 5.49 percentage points. Its accuracy is higher than that of ZARA with Gemini on seven of the eight datasets. The differences are particularly large on WISDM (65.6\% versus 83.68\%) and USCHAD (60.0\% versus 69.17\%); on Opportunity, \method\ reaches 85.00\%, compared with 92.5\% for ZARA.

Mean accuracy across the four \method\ configurations ranges from 80.83\% to 86.79\%. The GPT, DeepSeek, and Qwen configurations achieve 80.83\%, 83.48\%, and 83.52\%, respectively. These results indicate that hierarchical candidates and local evidence can be used with different foundation models, although final accuracy still depends on the model. The larger gain on WISDM is consistent with the design of first narrowing candidates and then loading activity-pair knowledge; the ablations below further assess these components. Per-activity accuracies are shown in Appendix Fig.~\ref{fig:per-activity-accuracy}.

\subsection{Cross-Dataset Experiments}
\begin{table}[!t]
\centering
\caption{Shared activities and sensor-position mappings in cross-dataset experiments.}
\label{tab:cross-dataset-mapping}
\rmfamily
\resizebox{\columnwidth}{!}{%
\begin{tabular}{ccc}
\toprule
Dataset pair & Activities & Sensor-position mapping\\
\midrule
UCIHAR $\leftrightarrow$ USCHAD & 5 & Waist $\leftrightarrow$ front right hip\\
\midrule
PAMAP2 $\leftrightarrow$ WISDM & 5 & Wrist $\leftrightarrow$ hand\\
\midrule
Mhealth $\leftrightarrow$ WISDM & 5 & Right wrist $\leftrightarrow$ hand\\
\midrule
\multirow{3}{*}{PAMAP2 $\leftrightarrow$ Mhealth} & \multirow{3}{*}{7} & Wrist $\leftrightarrow$ right wrist\\
 & & Chest $\leftrightarrow$ chest\\
 & & Ankle $\leftrightarrow$ left ankle\\
\bottomrule
\end{tabular}
}
\end{table}
Table~\ref{tab:transfer} reports eight transfer directions across four dataset pairs. The full method achieves mean accuracy of 88.81\% and macro-F1 of 88.74\%. Removing transferred feature knowledge yields 86.60\% mean accuracy, a difference of 2.21 percentage points. The full method has higher accuracy in six directions; for PAMAP2$\rightarrow$Mhealth, it reaches 97.14\%, compared with 90.71\% without transferred feature knowledge.

In cross-dataset experiments, the full method has 1.681 final candidates on average, with 69.55\% mean candidate reduction and 98.24\% mean true-activity retention; retention ranges from 94.00\% to 100.00\% across directions. Online retrieval still uses target-domain reference examples. The comparison without transferred feature knowledge retains the source-domain hierarchy but removes source-domain activity-pair features, isolating the contribution of that transferred knowledge.

\begin{table*}[!t]
\centering
\caption{Cross-dataset transfer results and accuracy without transferred feature knowledge. Candidate-set metrics are aggregated over query windows.}
\label{tab:transfer}
\rmfamily
\resizebox{\textwidth}{!}{%
\begin{tabular}{lrrrrrr}
\toprule
Transfer direction & \shortstack{Accuracy\\(\%)} & \shortstack{Macro-F1\\(\%)} & \shortstack{Accuracy without\\transferred features (\%)} & \shortstack{Mean final\\candidates} & \shortstack{Candidate\\reduction (\%)} & \shortstack{True-activity\\retention (\%)}\\
\midrule
UCIHAR$\rightarrow$USCHAD & 94.00 & 94.00 & 92.00 & 2.270 & 54.60 & 100.00\\
USCHAD$\rightarrow$UCIHAR & 93.00 & 92.90 & 94.00 & 1.130 & 77.40 & 97.00\\
WISDM$\rightarrow$PAMAP2 & 78.00 & 78.25 & 77.00 & 1.910 & 61.80 & 94.00\\
PAMAP2$\rightarrow$WISDM & 92.00 & 91.99 & 88.67 & 1.313 & 73.73 & 98.67\\
WISDM$\rightarrow$Mhealth & 75.00 & 74.00 & 71.00 & 1.780 & 64.40 & 100.00\\
Mhealth$\rightarrow$WISDM & 91.33 & 91.38 & 88.67 & 1.400 & 72.00 & 98.00\\
PAMAP2$\rightarrow$Mhealth & 97.14 & 97.13 & 90.71 & 1.850 & 73.57 & 100.00\\
Mhealth$\rightarrow$PAMAP2 & 90.00 & 90.27 & 90.71 & 2.336 & 66.63 & 98.57\\
\midrule
Arithmetic mean & \best{88.81} & \best{88.74} & \best{86.60} & \best{1.681} & \best{69.55} & \best{98.24}\\
\bottomrule
\end{tabular}}
\end{table*}

\subsection{Ablation Results and Token Cost}
We compare the full method and ablated variants in terms of accuracy and tokens per sample. Accuracy tests whether each component improves final classification; token count measures changes in interface input and output cost due to candidate reduction. The variant without the hierarchical structure does not retain candidates level by level. Instead, it sends all activities to the final decision while retaining retrieval evidence and registered activity-pair knowledge.

\subsubsection{Accuracy}
Table~\ref{tab:ablation} reports accuracy. The full method averages 86.79\% across the eight datasets, exceeding the variant without feature knowledge by 3.31 percentage points, the fixed-rule variant without an LLM by 3.84 points, and the variant without the hierarchical structure by 1.40 points. These comparisons assess local feature knowledge, LLM candidate selection, and the hierarchical inference procedure, respectively. The hierarchical procedure yields both an accuracy gain and token savings.

\begin{table}[!t]
\centering
\caption{Cross-subject accuracy (\%). Highest values are bold; ties are marked equally.}
\label{tab:ablation}
\rmfamily
\resizebox{\columnwidth}{!}{%
\begin{tabular}{lcccc}
\toprule
Dataset & Full & No features & No LLM & No hierarchy\\
\midrule
DSADS & \best{89.47} & 85.79 & 84.74 & 87.37\\
Mhealth & \best{88.75} & 87.08 & 86.67 & \best{88.75}\\
Opportunity & \best{85.00} & 81.00 & 80.00 & \best{85.00}\\
PAMAP2 & \best{82.50} & 80.83 & 80.42 & 81.67\\
Shoaib & \best{99.05} & \best{99.05} & \best{99.05} & \best{99.05}\\
UCIHAR & \best{96.67} & 95.00 & 92.08 & 96.25\\
USCHAD & \best{69.17} & 57.50 & 62.50 & 61.67\\
WISDM & \best{83.68} & 81.60 & 78.13 & 83.33\\
\midrule
Arithmetic mean & \best{86.79} & 83.48 & 82.95 & 85.39\\
\bottomrule
\end{tabular}}
\end{table}

\subsubsection{Token Use}
Table~\ref{tab:token} reports tokens per sample for the full method and the variants without feature knowledge or the hierarchical structure. Relative to the latter, the full method reduces total input and output tokens on all eight datasets. Savings range from 78.75\% to 90.68\%, with an arithmetic mean of 85.15\%. The full method uses more tokens than the variant without features because it reads activity-pair reports after candidate reduction. This additional input accompanies the mean accuracy gain of 3.31 percentage points in Table~\ref{tab:ablation}. Token use for the full method ranges from 1,013.4 per sample on Shoaib to 12,143.1 on WISDM, indicating that cost depends not only on the final candidate count but also on the number of registered pairs, available feature reports, and routing rounds.

\begin{table}[!t]
\centering
\caption{Tokens per sample and the full method's savings relative to the variant without the hierarchical structure.}
\label{tab:token}
\rmfamily
\resizebox{\columnwidth}{!}{%
\begin{tabular}{lrrrr}
\toprule
Dataset & Full & No features & No hierarchy & Savings (\%)\\
\midrule
DSADS & 4,679.3 & 935.0 & 44,674.6 & 89.53\\
Mhealth & 1,910.9 & 1,409.3 & 9,238.3 & 79.32\\
Opportunity & 3,111.1 & 1,104.6 & 14,642.6 & 78.75\\
PAMAP2 & 10,038.2 & 1,141.6 & 63,655.0 & 84.23\\
Shoaib & 1,013.4 & 787.4 & 8,484.7 & 88.06\\
UCIHAR & 1,803.5 & 814.9 & 10,666.5 & 83.09\\
USCHAD & 7,879.8 & 1,059.8 & 63,146.4 & 87.52\\
WISDM & 12,143.1 & 1,093.9 & 130,267.7 & 90.68\\
\midrule
Arithmetic mean & -- & -- & -- & \best{85.15}\\
\bottomrule
\end{tabular}}
\end{table}

\begin{table*}[!t]
\centering
\caption{Hierarchical candidate reduction. Root compression is calculated from the number of root activities and root candidate units; final reduction and true-activity retention are reported together.}
\label{tab:reduction}
\rmfamily
\resizebox{\textwidth}{!}{%
\begin{tabular}{lrrrrrr}
\toprule
Dataset & Root activities & Root units & Root compression (\%) & Final candidates & Final reduction (\%) & True-activity retention (\%)\\
\midrule
DSADS & 19.00 & 6.00 & 68.42 & 3.09 & 83.71 & 99.47\\
Mhealth & 12.00 & 8.00 & 33.33 & 3.01 & 74.93 & 100.00\\
Opportunity & 4.00 & 2.00 & 50.00 & 1.55 & 61.38 & 93.00\\
PAMAP2 & 12.00 & 6.00 & 50.00 & 2.65 & 77.95 & 92.50\\
Shoaib & 7.00 & 5.00 & 28.57 & 1.77 & 74.69 & 99.52\\
UCIHAR & 6.00 & 2.00 & 66.67 & 1.80 & 70.00 & 98.33\\
USCHAD & 12.00 & 2.00 & 83.33 & 2.35 & 80.42 & 90.42\\
WISDM & 18.00 & 5.00 & 72.22 & 2.85 & 84.18 & 98.26\\
\midrule
Arithmetic mean & -- & -- & -- & \best{2.38} & \best{75.91} & \best{96.44}\\
\bottomrule
\end{tabular}}

\end{table*}

\subsection{Does the Hierarchy Retain the True Activity?}
Table~\ref{tab:reduction} reports candidate-set size alongside true-activity retention. Across eight datasets, the final set contains 2.38 candidates on average, with 75.91\% mean reduction and 96.44\% true-activity retention. These metrics must be read together: a smaller set is useful only if it still contains the true activity. On DSADS, 19 activities are reduced to 3.09 candidates on average, with 99.47\% retention. On USCHAD, the set shrinks to 2.35 candidates, but retention is 90.42\%, indicating that some subsequent performance loss originates when upper-level routing removes the true activity. This diagnosis separates candidate-reduction errors from errors in final label selection.

\subsection{Local Processing Latency}
Table~\ref{tab:latency} reports mean time spent on retrieval, activity-pair report generation, and query feature extraction in the full method. Each operation takes less than one second per sample on every dataset; retrieval takes 0.005--0.043 seconds per sample. Report generation varies with the number of registered pairs among current candidates, whereas feature extraction varies with sensor positions and available modalities.
\begin{table}[!htbp]
\centering
\caption{Local processing latency of the full method (seconds per sample). Timings for individual components may overlap slightly.}
\label{tab:latency}
\rmfamily
\begin{tabular}{lrrr}
\toprule
Dataset & Retrieval & Pair reports & Query features\\
\midrule
DSADS & 0.043 & 0.369 & 0.150\\
Mhealth & 0.012 & 0.102 & 0.076\\
Opportunity & 0.018 & 0.510 & 0.160\\
PAMAP2 & 0.016 & 0.270 & 0.097\\
Shoaib & 0.019 & 0.244 & 0.106\\
UCIHAR & 0.005 & 0.045 & 0.036\\
USCHAD & 0.007 & 0.067 & 0.041\\
WISDM & 0.010 & 0.070 & 0.036\\
\bottomrule
\end{tabular}
\end{table}

\section{Conclusion}
We proposed \method, which uses stable source-domain activity confusions to determine which activities should be compared first. Variable-size candidate retention then narrows the set level by level, and validated sensor-feature knowledge is loaded only for registered pairs that remain. Across eight datasets, the full method achieves 86.79\% mean accuracy, with 2.38 final candidates, 75.91\% candidate reduction, and 96.44\% true-activity retention. It uses 85.15\% fewer tokens on average than the variant without the hierarchical structure. Across eight cross-dataset directions, mean accuracy is 88.81\%, compared with 86.60\% without transferred feature knowledge. These results support a clear principle for organizing inference: use reproducible confusion evidence to narrow the comparison scope before an LLM reads local knowledge directly relevant to the current candidates.
\appendix
\makeatletter
\@addtoreset{table}{section}
\@addtoreset{figure}{section}
\@addtoreset{equation}{section}
\makeatother
\renewcommand{\thetable}{\thesection.\arabic{table}}
\renewcommand{\thefigure}{\thesection.\arabic{figure}}
\renewcommand{\theequation}{\thesection.\arabic{equation}}
\setcounter{table}{0}
\setcounter{figure}{0}
\setcounter{equation}{0}
\section{Dataset Descriptions}
The experiments use eight public HAR datasets. Table~\ref{tab:appendix-datasets} summarizes their numbers of activities, sensor positions, and activity labels. Retrieval, hierarchy construction, and inference are performed within each dataset's label set.

\begin{table*}[!t]
\centering
\caption{Experimental datasets, sensor positions, and activity labels.}
\label{tab:appendix-datasets}
\rmfamily
{\scriptsize
\begin{tabular}{@{}p{0.11\textwidth}rp{0.20\textwidth}p{0.53\textwidth}@{}}
\toprule
Dataset & Activities & Sensor positions & Activity labels\\
\midrule
Opportunity & 4 & Back, upper arms, forearms & Standing, walking, sitting, lying\\
UCIHAR & 6 & Waist & Standing, sitting, lying, walking, descending stairs, ascending stairs\\
Shoaib & 7 & Left and right pockets, belt, right upper arm, right wrist & Walking, standing, jogging, sitting, cycling, descending stairs, ascending stairs\\
PAMAP2 & 12 & Wrist, chest, ankle & Lying, sitting, standing, ironing, vacuum cleaning, ascending stairs, descending stairs, walking, Nordic walking, cycling, running, rope jumping\\
USCHAD & 12 & Front right hip & Sleeping, sitting, elevator down, elevator up, standing, jumping, descending stairs, walking right, walking forward, running forward, ascending stairs, walking left\\
Mhealth & 12 & Chest, right wrist, left ankle & Ascending stairs, standing still, sitting relaxed, lying, walking, bending forward, lifting arms forward, bending knees, jogging, running, jumping forward and backward, cycling\\
WISDM & 18 & Hand & Walking, jogging, stairs, sitting, standing, typing, brushing teeth, eating soup, eating chips, eating pasta, eating a sandwich, kicking a ball, catching a ball, drinking, dribbling, writing, clapping, folding clothes\\
DSADS & 19 & Torso, left and right arms, left and right legs & Sitting, standing, lying supine, lying on right side, ascending stairs, descending stairs, standing in an elevator, moving in an elevator, walking slowly, rowing, jumping, treadmill walking on level ground, treadmill walking on an incline, fast treadmill running, stepping, cross-training, basketball, and two exercise-bicycle postures\\
\bottomrule
\end{tabular}
}
\end{table*}

\begin{table*}[!t]
\centering
\renewcommand{\thetable}{D.1}
\caption{Predefined features used for activity-pair knowledge.}
\label{tab:predefined-features}
\rmfamily
{\small
\begin{tabular*}{\textwidth}{@{}p{0.20\textwidth}@{\hspace{0.025\textwidth}}p{0.775\textwidth}@{}}
\toprule
Feature family & Representative statistics or derived quantities\\
\midrule
Time-domain statistics & Mean, standard deviation, variance, maximum, minimum, median, root mean square, peak, range, interquartile range, skewness, kurtosis, and signal magnitude area\\
Frequency-domain statistics & Low-, mid-, and high-frequency band powers and their ratios; dominant frequency and power; secondary peak frequency; spectral centroid, entropy, skewness, and kurtosis; weighted mean frequency; and spectral energy\\
Short-time Fourier transform & Maximum, mean, and standard deviation in each frequency band, plus statistics of short-time spectral entropy and spectral centroid\\
Change and impact & Mean, root mean square, and standard deviation of first differences; zero-crossing rate; impact rate of change; and peaks\\
Periodicity and dependence & First autocorrelation peak, first minimum, and first zero crossing; fourth-order autoregressive coefficients; and recurrence rate\\
Multiscale complexity & Wavelet-packet energy and permutation entropy\\
Posture and inter-axis relations & Gravity-direction statistics and inter-axis Pearson correlation coefficients\\
\bottomrule
\end{tabular*}
}
\end{table*}

\begingroup\small
\section{Dataset Preprocessing}
UCIHAR uses its publicly provided windows, and DSADS uses its predefined windows. All other signals are divided into nonoverlapping windows.
\begin{itemize}
\setlength{\itemsep}{1pt}\setlength{\parsep}{0pt}\setlength{\topsep}{3pt}
  \item \textbf{Opportunity:} 30 Hz; subject 4 provides queries and the other subjects provide reference data; 2-second windows; 50 query windows per class.
  \item \textbf{UCIHAR:} 50 Hz; the public training/test split is retained; 2.56-second windows with 50\% overlap; 40 windows per class are sampled from test subjects.
  \item \textbf{Shoaib:} 50 Hz; subjects 1 and 9 provide queries; 2-second windows; 30 windows per class.
  \item \textbf{PAMAP2:} 100 Hz; subjects 5 and 6 provide queries; 2-second windows; 20 windows per class.
  \item \textbf{USCHAD:} 100 Hz; subjects 13 and 14 provide queries; 2-second windows; 20 windows per class.
  \item \textbf{Mhealth:} 50 Hz; subjects 1 and 6 provide queries; 2-second windows; 20 windows per class.
  \item \textbf{WISDM:} The 20-Hz smartwatch hand subset is used. Accelerometer and gyroscope readings are aligned by timestamp. Forty-seven subjects without alignment anomalies are retained, eight of them for queries. Ten-second windows are used, with 16 windows per class.
  \item \textbf{DSADS:} 25 Hz; subjects 2 and 4 provide queries; the dataset's 5-second windows are used, with 10 windows per class.
\end{itemize}

\section{Cross-Dataset Preprocessing}
Cross-dataset experiments retain shared activities and map sensor positions according to Table~\ref{tab:cross-dataset-mapping}. Only the hierarchy and activity-pair feature knowledge are transferred from the source domain; queries always retrieve from the target-domain reference set and never access raw source-domain windows. The four bidirectional dataset pairs are processed as follows:
\begin{itemize}
\setlength{\itemsep}{1pt}\setlength{\parsep}{0pt}\setlength{\topsep}{3pt}
  \item \textbf{UCIHAR and USCHAD:} Walking, ascending stairs, descending stairs, sitting, and standing are aligned. USCHAD is downsampled to 50 Hz; 40 and 20 windows per class are sampled from the respective datasets.
  \item \textbf{PAMAP2 and WISDM:} Walking, standing, jogging, sitting, and stairs are aligned. PAMAP2 is downsampled to 20 Hz; 20 and 30 windows per class are sampled, respectively.
  \item \textbf{Mhealth and WISDM:} Walking, standing, jogging, sitting, and stairs are aligned. Mhealth is downsampled to 20 Hz; 20 and 30 windows per class are sampled, respectively.
  \item \textbf{PAMAP2 and Mhealth:} Lying, sitting, standing, walking, running, cycling, and ascending stairs are aligned. PAMAP2 is downsampled to 50 Hz; 20 windows per class are sampled from each dataset.
\end{itemize}
Sampling is balanced across activities, and no window is sampled twice within a dataset.
\endgroup

\section{Predefined Features}
Table~\ref{tab:predefined-features} groups the candidate features used for activity-pair knowledge by computation type. Features are computed from the corresponding sensor axes and magnitude sequences. Activity-pair models then select the feature families to load using cross-subject permutation importance and independent source validation.

\onecolumn
\section{Per-Class Metrics}
\begin{center}
\refstepcounter{figure}
\includegraphics[width=\textwidth,trim=18 18 18 18,clip]{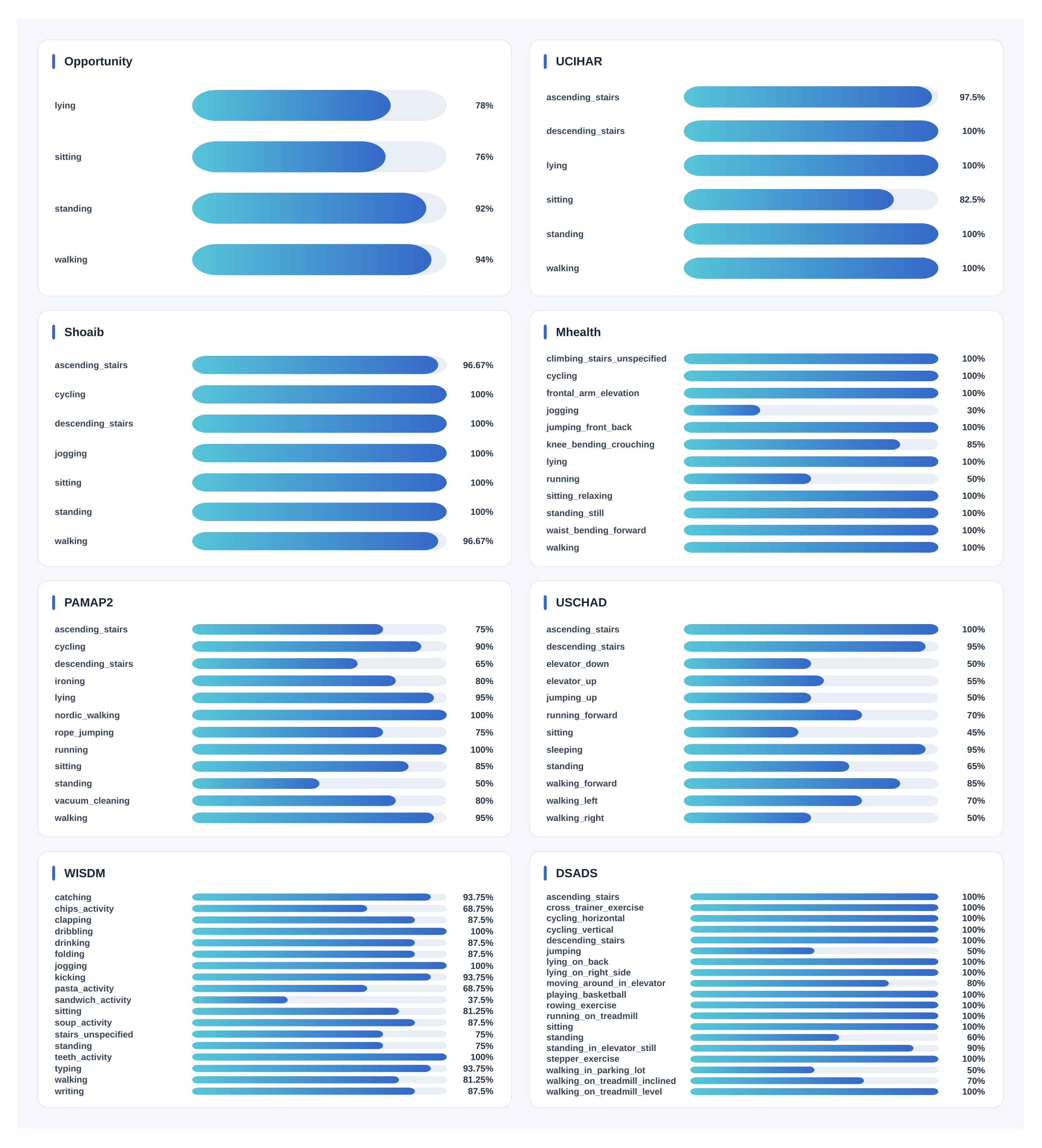}
{\small Fig.~\thefigure: Per-activity accuracy of the full method across eight datasets.\label{fig:per-activity-accuracy}\par}
\end{center}
\twocolumn
\section*{CRediT authorship contribution statement}
\noindent\textbf{Xiangqing Lu:} Conceptualization, Data curation, Investigation, Methodology, Visualization, Writing -- original draft.\par
\noindent\textbf{Tao Zhu:} Funding acquisition, Resources, Project administration, Supervision, Validation.\par
\noindent\textbf{Mingxing Nie:} Project administration, Supervision, Validation.\par
\noindent\textbf{Liming Chen:} Resources, Supervision, Validation.\par

\end{document}